\documentclass{article}
\usepackage{url}
\usepackage{graphicx}

\usepackage[utf8]{inputenc} % allow utf-8 input
\usepackage[T1]{fontenc}    % use 8-bit T1 fonts
\usepackage{booktabs}       % professional-quality tables
\usepackage{amsfonts}       % blackboard math symbols
\usepackage{nicefrac}       % compact symbols for 1/2, etc.
\usepackage{microtype}      % microtypography
\usepackage{float}
\usepackage{subcaption}
\usepackage{amsmath}
\usepackage{caption}
\usepackage{subcaption}

\usepackage[verbose=true,letterpaper]{geometry}
\AtBeginDocument{
  \newgeometry{
    textheight=9in,
    textwidth=5.5in,
    top=1in,
    headheight=12pt,
    headsep=25pt,
    footskip=30pt
  }
}

\begin{document}
\title{Linear Memory Networks}
\author{
	Davide Bacciu$^1$ \\\texttt{bacciu@di.unipi.it} \and
	Antonio Carta$^1$ \\ \texttt{antonio.carta@di.unipi.it}\and
	Alessandro Sperduti$^2$ \\ \texttt{sperduti@math.unipd.it}
}
\date{
    $^1$Dipartimento di Informatica, University of Pisa\\%
    $^2$Dipartimento di Matematica, University of Padova\\%
}

\maketitle

\begin{abstract}
Recurrent neural networks can learn complex transduction problems that require maintaining and actively exploiting a memory of their inputs. Such models traditionally consider memory and input-output functionalities indissolubly entangled. We introduce a novel recurrent architecture based on the conceptual separation between the functional input-output transformation and the memory mechanism, showing how they can be implemented through different neural components.  By building on such conceptualization, we introduce the Linear Memory Network, a recurrent model comprising a feedforward neural network, realizing the non-linear functional transformation, and a linear autoencoder for sequences, implementing the memory component. The resulting architecture can be efficiently trained by building on closed-form solutions to linear optimization problems. Further, by exploiting equivalence results between feedforward and recurrent neural networks we devise a pretraining schema for the proposed architecture. Experiments on polyphonic music datasets show competitive results against gated recurrent networks and other state of the art models.
\end{abstract}

\section{Introduction}
Recurrent Neural Networks (RNNs) are one of the pillars of the deep learning revolution, thanks to their statefulness which allows to learn complex computational tasks requiring the ability to memorize and "reason over" past inputs activations, such as with sequential data processing.  

The diffusion of RNN architectures has initially been restrained by the well known difficulties in learning long-term sequential dependencies due to gradient vanishing and explosion issues \cite{Bengio1994LearningLD}. Nonetheless, in the recent past, these issues have been addressed by a number of solutions exploiting gating units to control access and update of the state component, such as in the seminal LSTM model \cite{Hochreiter1997LongSM} and the follow-up GRU networks \cite{Chung2014EmpiricalEO}. A different line of research has tried tackling with the problem by resorting to articulated modular architectures, reducing the distance between long-term dependencies and introducing explicit multiscale time dynamics, such as in Clockwork RNN \cite{Koutnk2014ACR} and Hierarchical multiscale RNNs \cite{Chung2016HierarchicalMR}. Alternatively, RNN have been augmented with attention mechanisms \cite{Bahdanau2014NeuralMT} in the attempt of optimizing the state encoding by allowing to focus only on past memories at certain timesteps that are deemed relevant for the task. Attention mechanisms typically come at the cost of an increased computational effort, motivating recent attempts to improve it through hierarchical approaches \cite{Andrychowicz2016LearningEA}.

The common thread running through the solutions proposed so far is that of resorting to complex architectures, either at the level of the memory cell, such as with the gating units in LSTM, or at the network level, such as with multi-scale RNN and attention-based models. This results in models that, even if end-to-end differentiable, are often difficult to train, in practice. 

We introduce a novel RNN paradigm, dubbed Linear Memory Networks (LMN), which aims at simplifying the design and training of RNNs while retaining the ability to learn long-term dependencies. The model is based on the intuition that, in order to efficiently solve a sequence processing problem, recurrent models need to solve two associated tasks: a functional task, which concerns mapping the input sequence into a sequence of outputs, and a memory task, exploiting a memorization mechanism to remember past states that can serve for the functional task \cite{Sperduti2015EquivalenceRB}. Recurrent models typically solve these two tasks together, by learning the mapping from inputs to outputs and the memorization mechanism at the same time. The LMN puts forward a novel approach based on the explicit separation between the functional and memory components of a recurrent model. The key intuition is that by explicitly separating the two tasks it is possible to simplify both the architecture and the learning algorithms used to train these models, while acquiring a deeper understanding of the inner workings, for example exploiting explicit memorization.

The literature reports several attempts to introduce a separate memory for recurrent architectures, such as in Memory Networks \cite{sukhbaatar2015end} and in Neural Turing Machines \cite{graves2014neural}.  Differentiable Neural Computers \cite{graves2016hybrid}. However, the memory mechanism in these models is used to augment architectures that are already recurrent, rather than to simplify them, and it typically involves non-trivial addressing and memory access schemes. The end result is that these architectures become easily quite complex and difficult to train.

The LMN, on the other hand, proposes a simple architecture comprising a non-linear feedforward network to model the functional component of the RNN, while the memory component is realized by means of a linear autoencoder for sequences. The choice of these components allows us to exploit closed-form solutions for linear autoencoder training by \cite{sperduti2013linear} and the equivalence results between certain classes of RNN and their unrolled feedforward version \cite{Sperduti2015EquivalenceRB} to efficiently train the linear memory to reconstruct the hidden states computed by the nonlinear feedforward part. We will show how this allows to define a simple multi-stage learning scheme, comprising an effective pretraining phase that cannot be realized in gated architectures, such as LSTM. Through an experimental analysis on complex sequence processing tasks, we show how the simple LMN architecture is capable of obtaining competitive results with respect to complex recurrent models, including gated RNN.

\section{Linear Autoencoder for Sequences}
\label{sec:la}
We begin by summarizing the linear autoencoder for sequences \cite{sperduti2013linear}, that is the building block for realizing the LMN memory component. A linear autoencoder for sequences is a recurrent linear model designed to encode an input sequence into an hidden state, computed using a linear transformation. Given a set of sequences of input vectors $\{s^q=x^q_1...x^q_{l_q} | q = 1 \hdots n, x^q_j \in \mathbb{R}^a \}$, a linear autoencoder computes the state vector $y_t \in \mathbb{R}^p$, i.e. the encoding of the input sequence up to time $t$, using the following equations:
\begin{equation} \label{eq:AE1}
y_t = Ax_t + By_{t-1}
\end{equation}
\begin{equation} \label{eq:AE2}
\begin{bmatrix} x_t \\ y_{t-1} \end{bmatrix} = Cy_t,
\end{equation}

where $A \in \mathbb{R}^{p \times a}$, $B \in \mathbb{R}^{p \times p}$ and $C \in \mathbb{R}^{(a+p) \times p}$ are the model parameters, which can be trained by exploiting a decomposition of the output data matrix $Y$. Let us assume that the training set consists of a single sequence $\{x_1, \hdots, x_l\}$ and define $Y \in \mathbb{R}^{l \times p}$ as the matrix containing the state vectors at each timestep. From Eq. (\ref{eq:AE1}) and (\ref{eq:AE2}) it follows that:

\begin{equation} 
\underbrace{
\begin{bmatrix}
y_1^T \\ y_2^T \\ y_3^T \\ \vdots \\ y_l^T
\end{bmatrix}}_{Y} = 
\underbrace{\begin{bmatrix}
x_1^T & 0 & \hdots & 0 \\
x_2^T & x_1^T & \hdots & 0 \\
\vdots & \vdots & \ddots & \vdots \\
x_l^T & x_{l-1}^T & \hdots & x_1^T
\end{bmatrix}}_{\Xi}
\underbrace{\begin{bmatrix}
A^T \\ A^T B^T \\ \vdots \\ A^T {B^{l-1}}^T
\end{bmatrix}}_{\Omega}.
\end{equation}

The matrix $\Xi$ contains the reversed subsequences of $x$ and $Y$ contains the state vectors at each timestep.
The encoding matrices $A$ and $B$ can be identified by exploiting the truncated SVD decomposition $\Xi = V \Sigma U^T$, where imposing $U^T \Omega = I$ yields $\Omega = U$. We can then exploit the structure of $\Xi$ to obtain $A$, $B$, and the associated matrix 
$C = \begin{bmatrix} A^T \\ B^T \end{bmatrix}$, as shown in \cite{sperduti2013linear}. Specifically, $\Omega=  U$
is satisfied by using matrices
\begin{equation*}
 P  \equiv  \left[\begin{array}{c} I_{a} \\ 0_{a(l - 1)\times a}\end{array}\right],\ \ \mbox{and}\ \ R  \equiv  \left[\begin{array}{ll} 0_{a\times a(l - 1)} &  0_{a\times a}\\ I_{a(l - 1)} & 0_{a(l - 1)\times a}\end{array}\right],
\end{equation*} 
to define
$A\equiv U^T P$ and \mbox{$B\equiv U^T  R U$}, where $I_{u}$ is the identity matrix of size $u$, and $0_{u\times v}$ is the zero matrix of size $u\times v$. 

The algorithm can be easily generalized to multiple sequences by stacking the data matrix $\Xi_q$ for each sequence $s^q$ and padding with zeros to match sequences length.

The sequence autoencoding scheme in Eq. (\ref{eq:AE1}) and (\ref{eq:AE2}) can be used to reconstruct the input sample and the past state given the current state vector. It should be clear how the iterative application of this process allows to reconstruct (an approximation of) the past input sequence. In particular,  the training algorithm guarantees an optimal encoding when $p = rank(\Xi)$. In the following sections, we show how such properties can be used to efficiently memorize and gather access to the history of the hidden states in a recurrent network. 

\section{Equivalence Results}
\begin{figure*}[!ht]
\centering
  \begin{subfigure}[t]{.4\textwidth}
    \centering
    \includegraphics[width=.95\textwidth]{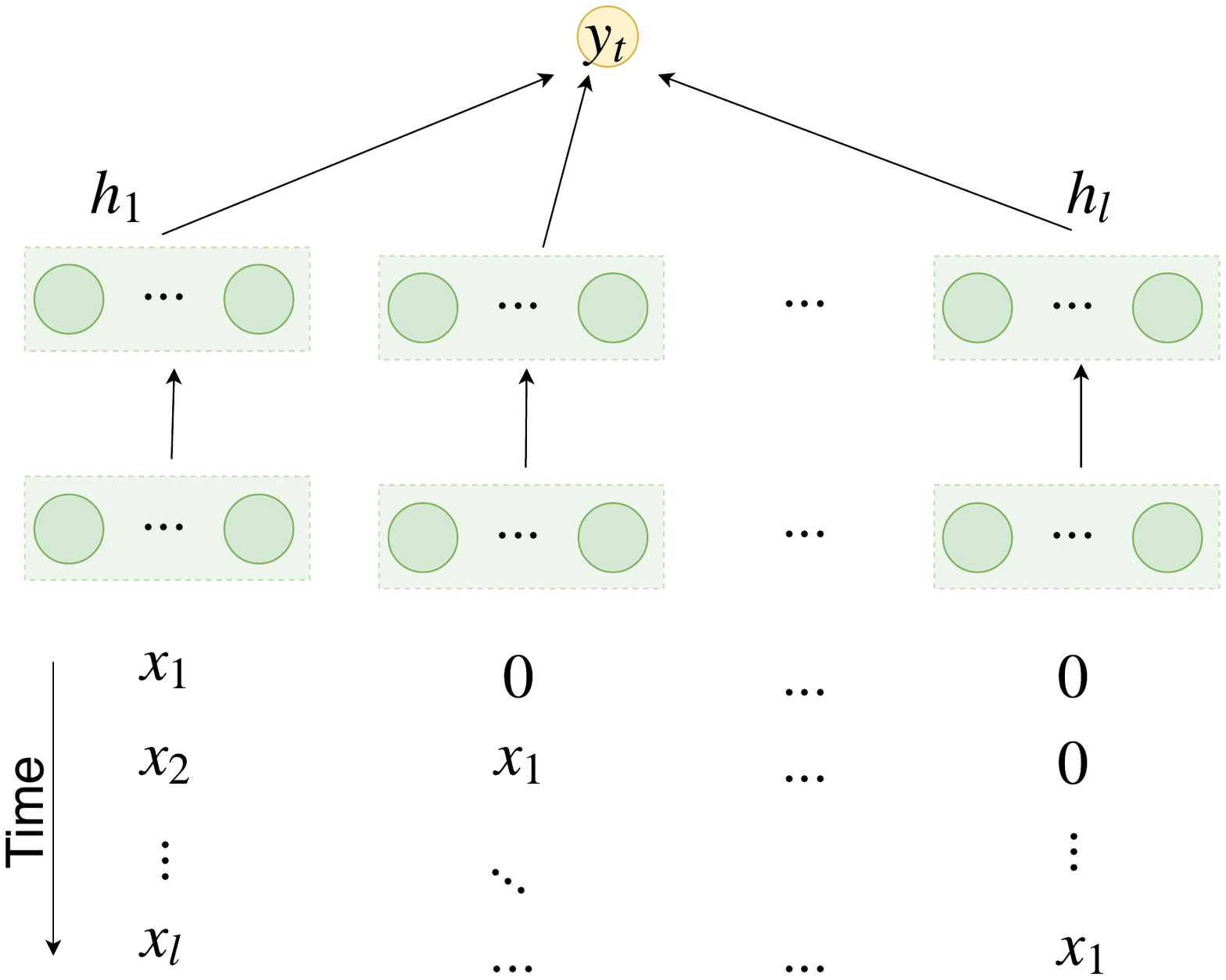}
    \caption{}
    \label{fig:fnn_net}
  \end{subfigure}%
  \begin{subfigure}[t]{.4\textwidth}
    \centering
    \includegraphics[width=.95\linewidth]{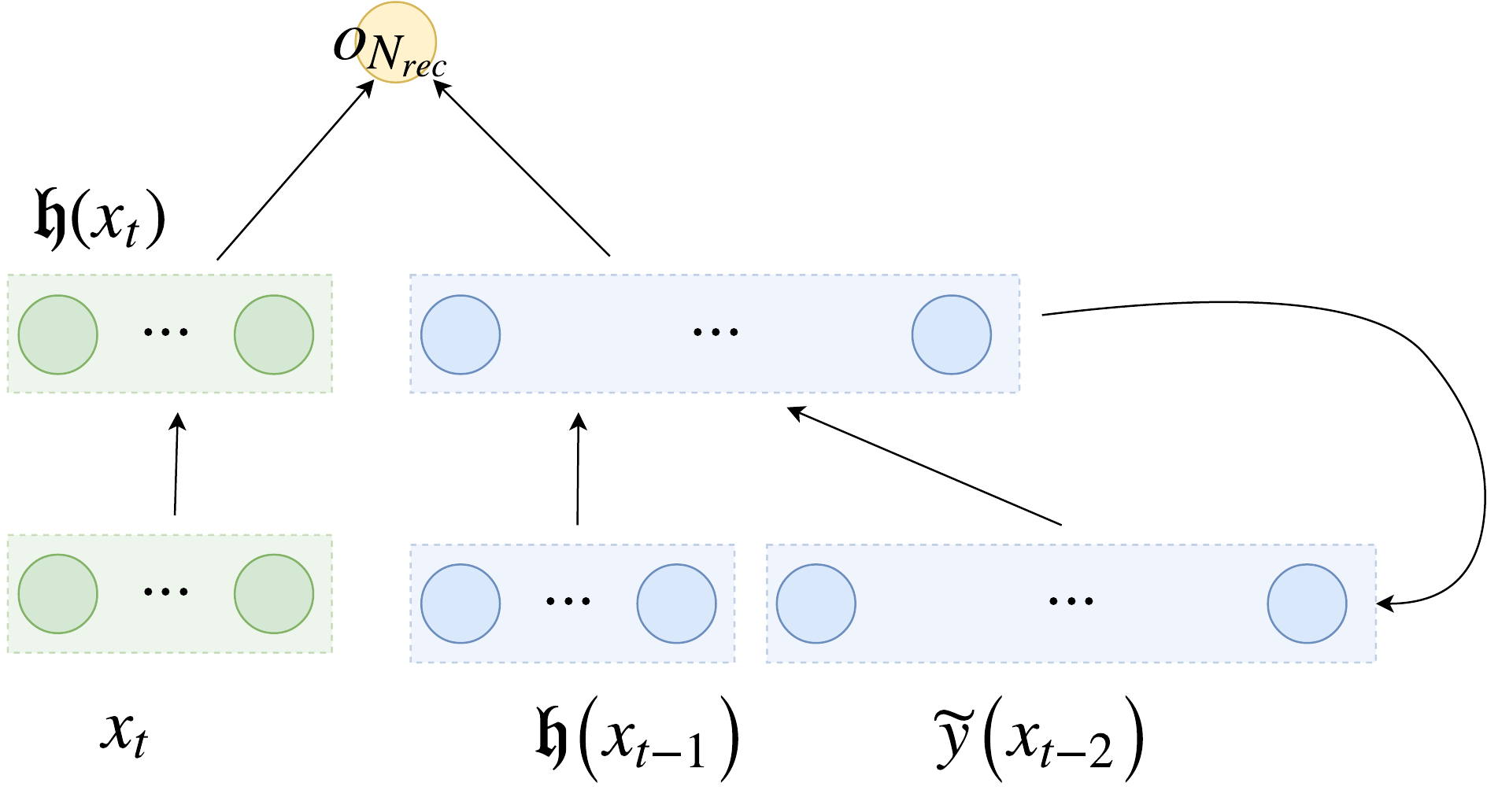}
    \caption{}
    \label{fig:rnn_net}
  \end{subfigure}
  \caption{FNN and equivalent RNN obtained by the equivalence result. Figure \ref{fig:fnn_net} shows the FNN, while  figure \ref{fig:rnn_net} shows the corresponding RNN, highlighting in blue the memorization units computed by the linear autoencoder.}
\end{figure*}

In this section we review the main results in \cite{Sperduti2015EquivalenceRB}.
Consider a sequence $s = (x_1, \hdots,\ x_l)$ and a feedforward neural network (FNN) that takes as input the reversed subsequences of $s$ at each timestep (as shown in Figure \ref{fig:fnn_net}). 

We restrict our discussion to FNN with a single hidden layer, single output unit, and the 1-time-step property, meaning that any hidden unit can connect only input units from the same timestep. Given the vectorial representation $x$ of a subsequence, e.g. $x=[x_3,x_2,x_1,0,\ldots,0]^\top$ for subsequence $(x_1, x_2, x_3)$, the output of the FNN can be computed as:
\begin{equation}
o_{N_f}(x) = \sigma\Big(\sum_{t=1}^{l}\sum_{h=1}^{H_h} w_{ht}^o \sigma(w_{ht}^\top x_t)\Big)
\end{equation}
where $H_i$ is the number of hidden units connected to the input element $x_i$, $N_H = \sum_{i=1}^tH_i$. We are interested in building an equivalent RNN, defined as:
\begin{eqnarray}
o_{N_{rec}}(x_t) &=& \sigma(c^\top y_t) \\
y_t &=& \sigma(Ax_t + By_{t-1}).
\end{eqnarray}
% aggiungere figura con FNN 1-timestep e corrispondente RNN
Consider the vector $\mathfrak{h}(x_i)$ collecting all the hidden contributions of the element $x_i$ at each timestep:
\begin{equation}
\mathfrak{h}(x_i) = [h_{11}(x_i),\ h_{21}(x_i), \hdots,\ h_{H_1 1}(x_i),\ h_{12}(x_i), \hdots,\ h_{H_l l}(x_i)].
\end{equation}

We can build the matrix ${\cal H}$ containing the reversed subsequences of $\tilde{s} = (\mathfrak{h}(x_1),\ \hdots,\ \mathfrak{h}(x_l))$:

\begin{equation}
{\cal H} = \begin{bmatrix}
\mathfrak{h}_1^T & 0 & \hdots & 0 \\
\mathfrak{h}_2^T & \mathfrak{h}_1^T & \hdots & 0 \\
\vdots & \vdots & \ddots & \vdots \\
\mathfrak{h}_l^T & \mathfrak{h}_{l-1}^T & \hdots & \mathfrak{h}_1^T
\end{bmatrix}.
\end{equation}

The matrix ${\cal H}$ contains all the information necessary to build an equivalent RNN.
We can compute the output $o_{N_f}(s)$ as:

\begin{eqnarray*}
\tilde{w} &=& [w_{1l}^o, \hdots,\ w_{H_l l}^o,\ \underbrace{0, \hdots,\ 0}_{N_H},\ w_{1 (l-1)}^0, \hdots,      \\
		& &\ w_{H_{l-1} (l-1)}^o, \underbrace{0, \hdots,\ 0}_{N_H}, \hdots,\ 0, w_{11}^o, \hdots,\ w_{H_1 l}^o]   \\
\mathfrak{h}_{j, t} &=& row\Big(\sum_{k=1}^{j-1} H_k + t, {\cal H}\Big) \\
o_{N_f}(s) &=& \sigma(\tilde{w} \mathfrak{h}_{j, t}^\top)
\end{eqnarray*}

where $row(i, A)$ takes the i-th row of matrix $A$. The RNN processes the elements $x_i$ one at a time but it can memorize the contributions of each $\mathfrak{h}(x_i)$ using a linear autoencoder. Let ${\cal H}_{[1, N_H]}$ be the first $N_H$ columns of the matrix ${\cal H}$, and ${\cal H}_{/[1,N_H]}$ the remaining columns. The equivalent RNN will have $N_H$ hidden units corresponding to $\mathfrak{h}_i$ and $\rho^{N_{/[1, N_H]}} = rank({\cal H}_{/[1, N_H]})$ hidden units that will act as the memory component and will compute the hidden state of the linear autoencoder trained on the matrix ${\cal H}_{/[1, N_H]}$.
Therefore, we can define $o_{N_{rec}}$ as:
\begin{eqnarray*}
o_{N_{rec}}(x_t) &=& \sigma(\tilde{w} \begin{bmatrix}
	I & 0  \\
    0 & U
\end{bmatrix} y_t) \\
y_t &=& \begin{bmatrix}
	\mathfrak{h}(x_t) \\
    \tilde{y}_{t-1}
\end{bmatrix} \\
\tilde{y}_t &=& \begin{bmatrix}
	A & B
\end{bmatrix} y_{t - 1}
\end{eqnarray*}
where $U$, $A$, $B$, are the corresponding matrices obtained by training the linear autoencoder on ${\cal H}_{/[1, N_H]}$. 

The corresponding RNN requires $N_H + \rho^{{\cal H}_{/[1, N_H]}}$ hidden units: a schematic view of its architecture is shown in Figure \ref{fig:rnn_net}. In \cite{Sperduti2015EquivalenceRB} are provided additional results that can reduce the number of hidden units required to construct the equivalent network. The resulting RNN can be seen as composed of two components: a functional component that computes the contribution of each element, corresponding to the computation of $\mathfrak{h}(x_i)$ in the first $N_H$ hidden units, and a memory component that memorize each contribution with a linear autoencoder. The Linear Memory Network, presented in the following section, is based on the same principles by making this separation explicit in the architecture. 

\section{Linear Memory Networks}
\begin{figure*}[!ht]
\centering
  \begin{subfigure}{.2\textwidth}
    \centering
    \includegraphics[width=.95\textwidth]{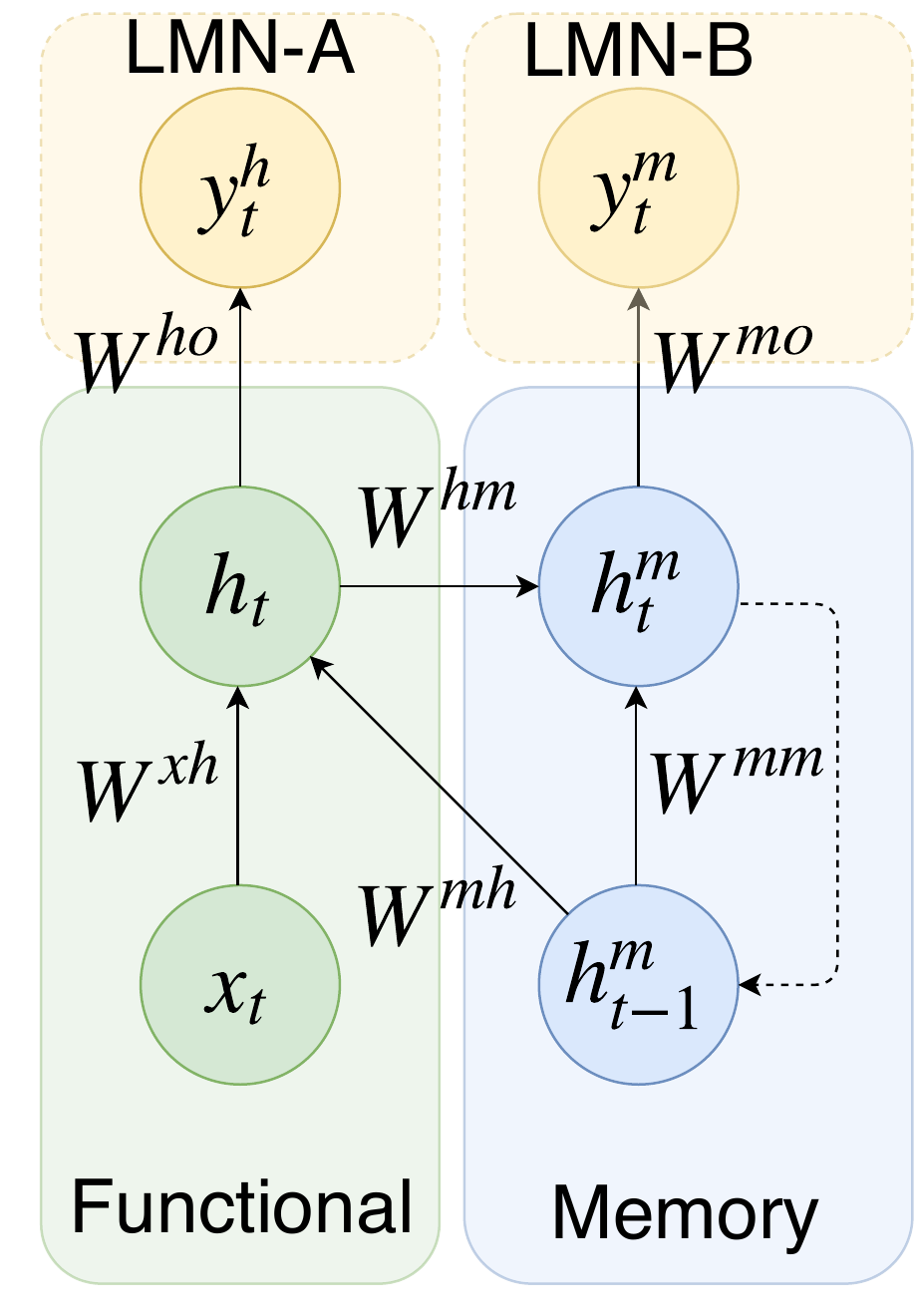}
    \caption{}
    \label{fig:lmn}
  \end{subfigure}%
  \begin{subfigure}{.37\textwidth}
    \centering
    \includegraphics[width=.95\linewidth]{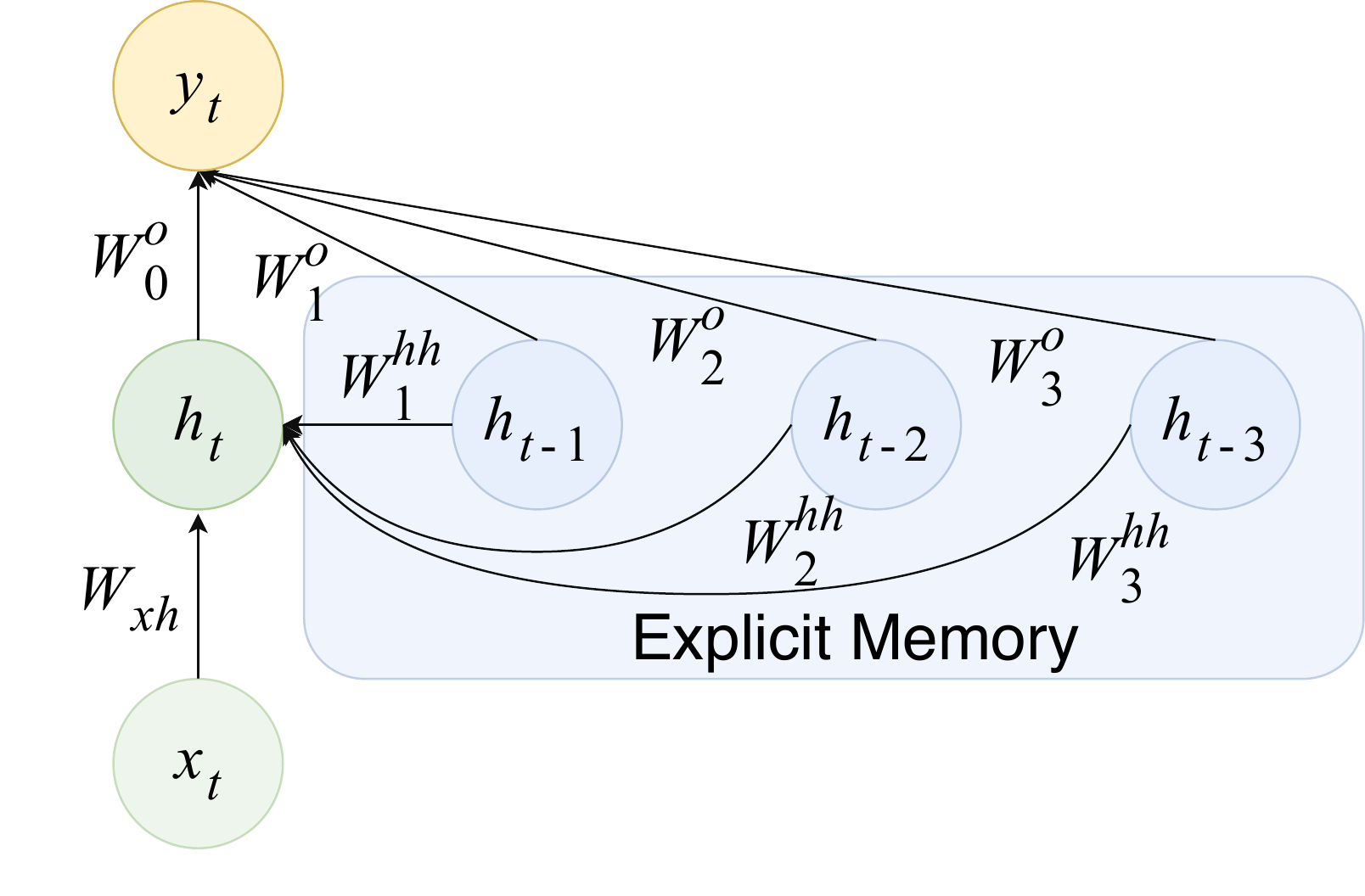}
    \caption{}
    \label{fig:ulm_memory}
  \end{subfigure}  
  \begin{subfigure}{.37\textwidth}
    \centering
    \includegraphics[width=.95\linewidth]{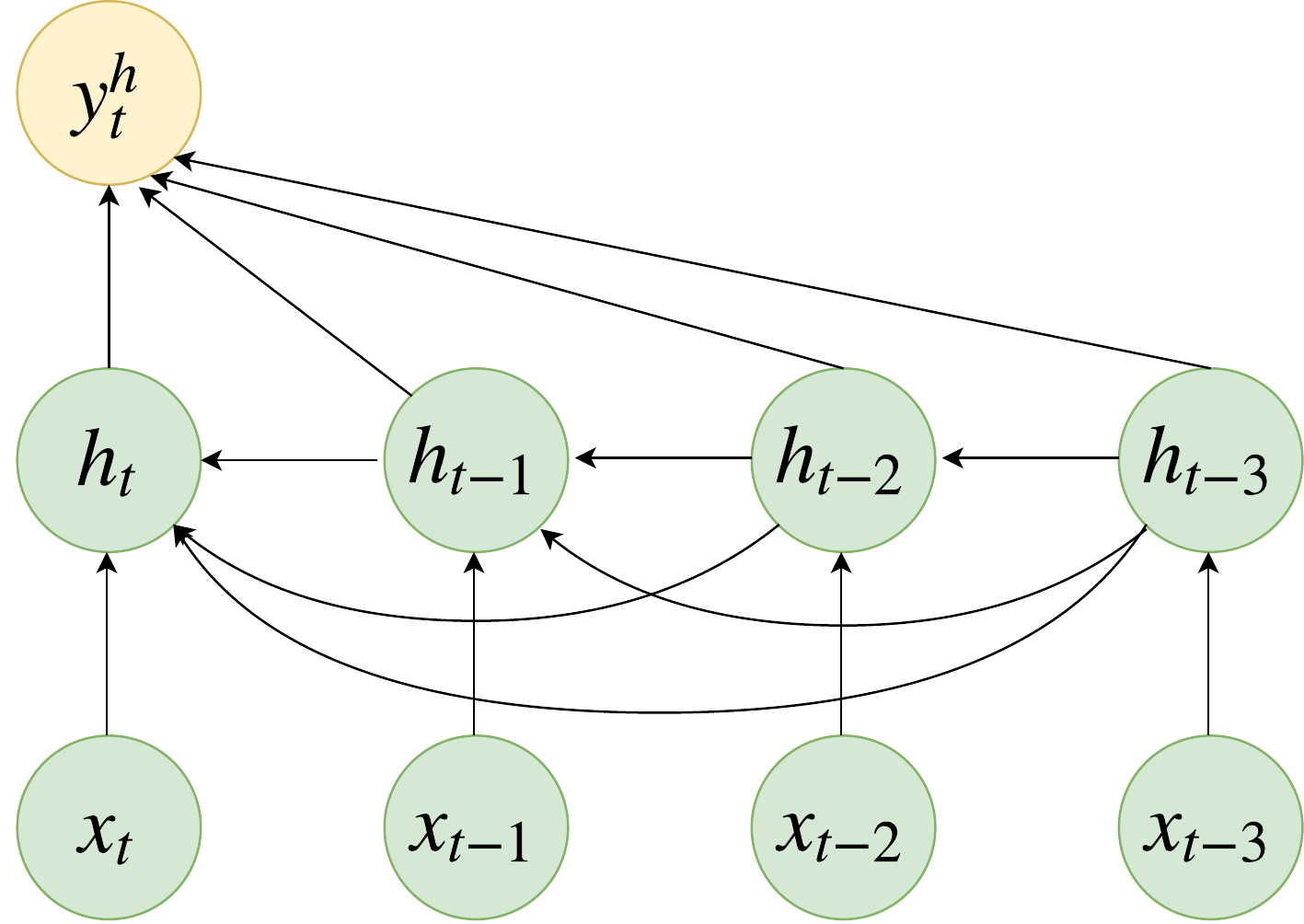}
    \caption{}
    \label{fig:ulm_unfold}
  \end{subfigure}
  \caption{Schematic representation of the memory layouts in the different network architectures. Figure \ref{fig:lmn} shows the architecture of the LMN, highlighting the separation between the functional and memory components and showing how the memory component is efficiently encoded using a linear autoencoder.  Figure \ref{fig:ulm_memory} shows the corresponding network with a (highly parameterized) explicit memory. Figure \ref{fig:ulm_unfold} shows the unfolded network for $k=3$ timesteps, where there is no explicit memory and, instead, we have explicit connections between hidden states.}
\end{figure*}

The Linear Memory Network (LMN) is a recurrent architecture where the memory and the functional components are explicitly separated: a sketch of the LMN structure is depicted in Figure \ref{fig:lmn}. The network combines a non-linear feedforward model (Functional box in Figure \ref{fig:lmn})  with a separate memory component implemented through a linear sequential autoencoder (Memory box in Figure \ref{fig:lmn}). Therefore, the memory is entirely linear while the feedforward component allows to model nonlinear dependencies between the input vectors. Note that the functional component comprises a number of feedforward neurons which is, in general, different from the number of recurrent linear units in the memory component.  The relationships between the functional activation $h_t\in \mathbb{R}^{p}$ and the  memory state $h^m_t\in \mathbb{R}^{m}$ are regulated by the following equations:

\begin{eqnarray}
h_t & = & \sigma(W^{xh}x_t + W^{mh}h^m_{t-1})  \label{lmn:1}\\
h^m_t & = & W^{hm}h_t + W^{mm}h^m_{t-1} \label{lmn:2}
\end{eqnarray}

\noindent where $a,\ p,\ m$ are respectively the input size, hidden size and memory size, while $W^{xh} \in \mathbb{R}^{p \times a}$, \mbox{$W^{mh} \in \mathbb{R}^{p \times m}$}, $W^{hm} \in \mathbb{R}^{m \times p}$, $W^{mm} \in \mathbb{R}^{m \times m}$ are the model parameters matrices, and $\sigma$ is a non-linear activation function (tanh for the purpose of this paper). The catch of the LMN architecture is using the linear autoencoder to linearly encode the history of the nonlinear functional 
activation $h_t$, i.e. the input to the autoencoder, in the state $h^m_t$.
The architecture is based on the equivalence results described in the previous section. The separation of the memory component allows to train the network to explicitly store the past activations of the functional component by training the corresponding linear autoencoder. A possible instantiation of this approach will be given in the next section, where we describe a pretraining scheme that exploits the properties of the network.

The network output (or a successive layer in a deeply layered architecture) can be wired to the LMN recurrent layer in two different ways, denoted as LMN-A and LMN-B in Figure \ref{fig:lmn}. The first approach, exploits the activation of the functional component $h_t$, while the second has direct access to the memory  $h^m_t$ , resulting in the following (alternative) output activations

\begin{eqnarray} 
y^m_t & = & \sigma(W^{mo}h^m_t) \label{lmn:3}\\
y^h_t & = & \sigma(W^{ho}h_t),\label{lmn:4}
\end{eqnarray}

where $\sigma$ is the activation function (sigmoid in this paper), $o$ the output size, $W^{mo} \in \mathbb{R}^{o \times m}$ and $W^{ho} \in \mathbb{R}^{o \times h}$ are the functional-to-output and memory-to-output parameter matrices, respectively.

We provide experimental results for both variants of the model. The definition given here and the experimental results cover only the case where the functional component is made of a single layer, but the approach can be easily extended to deep networks by adding layers to the feedforward component and connecting them with the memory. 

\subsection{Training and pretraining algorithm}
\label{sec:pretrain}
The LMN is differentiable end-to-end and it can be trained with standard backpropagation. If the memory component is trained using backpropagation there are no theoretical guarantees on its memorization properties. Therefore we are interested in alternative algorithms to train separately the memory of the network. Since the memory component is equivalent to a linear autoencoder, it can be trained separately, through an ad-hoc algorithm, to reconstruct the hidden representation of the functional component at previous time steps. By building on the explicit solution given in \cite{sperduti2013linear}, it is possible to construct an optimal encoding of the hidden states sequences with the minimal number of memory units. In this section, we propose a pretraining algorithm that can be used to initialize the model parameters based on the considerations above. 

The pretraining algorithm works in three steps. First, we construct through unfolding an equivalent network to generate an approximation of the functional component activations. Second, we use these activations as inputs to the linear autoencoder. Finally, we initialize the LMN by transferring to the LMN the output weights from the unfolded network, and the encoding weights from the linear autoencoder.   

More in detail, the first step of the pretraining algorithm constructs an unfolded version of the LMN, shown in Figure \ref{fig:ulm_unfold}, where the memory is substituted by an explicit representation of the previous hidden states and their relationship with the current state is explicitly represented by a parameterized transformation. To allow an efficient training of the network, the model is unrolled only for a fixed number of steps $k$. The unfolded model is trained to predict the network output $y_t^h$ and its  parameters are adjusted accordingly, using standard backpropagation. Each output is computed using only the last $k$ hidden state vectors. To predict the first $k$ output vectors we pad with zeros the missing inputs. The unrolled network is defined by the following equations:
\begin{eqnarray} 
h^t & = & \sigma(W^{xh}x^t + \sum_{i=1}^{k} W^{hh}_{i}h^{t-i}), \quad t=1, \hdots, k \label{ulm:1}\\
y^t & = & \sigma(\sum_{i=0}^{k} W^o_i h^{t-i}),\label{ulm:2}
\end{eqnarray}
where $W^{hh}_{i} \in \mathbb{R}^{h \times h}$ explicitly represents the relationship between the current hidden state and the hidden state at time $t-i$, while $W^o_i \in \mathbb{R}^{o \times h}$ represents the relationship between the current output $y^t$ and the hidden state at time $t-i$. 

The second step of the pretraining algorithm is based on previous equivalence results between recurrent and feedforward networks in   \cite{Sperduti2015EquivalenceRB}. While their results are focused on RNN, they can be easily adapted to the LMN architecture. Given a trained unfolded network as defined in Eq. (\ref{ulm:1}) and (\ref{ulm:2}), we want to create a new neural network with an explicit memory. Figure \ref{fig:ulm_memory} shows a network, equivalent to the unfolded one in Figure\ref{fig:ulm_unfold}, where previous hidden states are explicitly stored in a separate memory and used to compute the new hidden state. 

The explicit memory representation in Figure \ref{fig:ulm_memory} is inefficient because the computational cost and the number of parameters scale linearly with the memory size, which is explicitly bound by the finite length $k$. Instead, by using the compressed representation of the linear autoencoder in the LMN (see Figure \ref{fig:lmn}) we obtain a more efficient memory, since the hidden states are stored and compressed using a basis of principal components, as discussed in \cite{sperduti2013linear}. Note how such compressed representation does not depend on the original unfolding length $k$, whereas it allows to adaptively accommodate longer term dependencies up to the limits of memory capacity.

In essence, the second step of the pretraining algorithm amounts to training a linear autoencoder to reconstruct the hidden states $h^i$ (obtained from step one) for each sequence in the training set, by constructing the matrix $H$ of hidden state subsequences and using the linear autoencoder training algorithm. As a result, we obtain the matrices $A$ and $B$, corresponding to the parameters of the trained linear autoencoder. 

In the third step, we use the parameter matrices obtained in the previous two steps to initialize an LMN according to the following scheme. The LMN $W^{xh}$ matrix is initialized using the corresponding matrix $W^{xh}$ of the unfolded network. Then, given $U=[A^T, A^TB^T, \hdots, A^TB^{k-1}]$ obtained at step two, the remainder of the LMN matrices is initialized as
\begin{eqnarray}
W^{hm} & = & A \label{eq:pret}\\
W^{mm} & = & B \\
W^{mh} & = & \begin{bmatrix}W^{hh}_1 & \hdots & W^{hh}_k \end{bmatrix} U \\
W^{mo} & = & \begin{bmatrix} W^{o}_1 & \hdots & W^{o}_k \end{bmatrix} U.
\label{eq:pret2}
\end{eqnarray}

Using this procedure the memory is initialized to reconstruct the entire sequence of hidden states computed by the unrolled network. While the unrolled network requires a number of parameters that scales with the unrolling length $k$, the LMN is more efficient and can use the linear autoencoder to reduce the number of parameters without reducing the memory length.

\subsection{A comparison with gated recurrent architectures\label{comparison}}
The main difference between the LMN model and other recurrent architectures in the literature is the conceptual separation between the memory and the functional component. The state dynamics is captured by the linear memory component without the need for multiplicative gates like in LSTM and GRU units, leading to a simple, easily trainable architecture, without unwanted exponential decay effects due to the presence of gates. The number of model parameters, having fixed the number of neurons, is also smaller in LMN: LSTM requires $4(x + h)h$ parameters, GRU requires $3(x + h)h$ parameters, and LMN requires $(x + m)f + (f + m) m$ parameters, where $x$ is the input size, $h$ the number of hidden units, $f$ and $m$ the number of functional and memory units (only for LMN). If we set $h=m+f$, we obtain that the number of parameters for the LMN architecture is maximized when $m = f = h/2$. The total number of parameters for the LMN architecture in this case becomes $\frac{1}{2} ((x + \frac{h}{2})h + h^2)$, less than LSTM and GRU architectures with the same number of hidden units. Further, the linear dynamics of the memory allows the design of ad-hoc, optimized training algorithms. As an example of this possibility, this paper presents a pretraining algorithm.  Another interesting possibility is the development of second order optimization methods \cite{Martens2010DeepLV} which exploit the linearity of the memory to yield an efficient closed form solution. 

\section{Experimental Results} \label{sec:exp}
We evaluated LMN on sequence prediction tasks using four different datasets of polyphonic music representing piano roll versions of songs in different styles and with different degrees of polyphony~\cite{BoulangerLewandowski2012ModelingTD}. Each sequence is sampled at equal timesteps to obtain a feature vector composed of $88$ binary values representing the piano notes from A0 to C8. Each note is set to $1$ if it is currently being played or $0$ if it is not. The task is to predict the notes played at the next timestep given the sequence of previous notes. The performance of each model is evaluated using frame-level accuracy as defined in \cite{Bay2009EvaluationOM}. We used the same train-validation-test split as in \cite{BoulangerLewandowski2012ModelingTD}. Even if all datasets contain music represented in piano roll style, they are different from each other, ranging from classical music to folk music, composed for piano, orchestra or chorales. This generates widely different performance results depending on the dataset. Table \ref{tbl:midi-data} shows the number of samples and the maximum length of the sequences for each datasets.

\begin{table}
\centering
\caption{Number of sequences and maximum length for each dataset.}
\label{tbl:midi-data}
\begin{tabular}{lcc}
\toprule
              & Samples & max. timesteps \\
\toprule
JSB Chorales  & 382       & 160 \\
MuseData      & 783       & 4273 \\
Nottingham    & 1037      & 1793 \\
Piano MIDI    & 124       & 3857 \\
\bottomrule
\end{tabular}
\end{table}

We compare the output configurations of the LMN architecture in Figure \ref{fig:lmn} using a random initialization of the model parameters (LMN-A and LMN-B, in the following). In addition, we have tested the LMN-B output configuration with parameters initialized using the pretraining scheme (pretraining results are shown only for the LMN-B configuration as including those of pretrained LMN-A would not add much to the analysis). The LMN results are compared versus a number of reference models from literature. Specifically, we consider an RNN with random initialization or using the pretraining scheme described in \cite{Pasa2014PretrainingOR}, an LSTM network, and the RNN-RBM model (for which we report the original results from \cite{BoulangerLewandowski2012ModelingTD}). Note that the Nottingham dataset has been expanded since the publication of \cite{BoulangerLewandowski2012ModelingTD} and therefore the results are not fully comparable.
\begin{table*}[!ht]
\centering
\caption{Frame-level accuracy computed on the test set for each model. RNN-RBM results are taken from \cite{BoulangerLewandowski2012ModelingTD}}
\label{tbl:midi-results}
\begin{tabular}{lcccc}
\toprule
            & JSB Chorales & MuseData & Nottingham & Piano MIDI \\
\toprule
RNN		    & 31.00        & 35.02    & 72.29      & 26.52		\\
pret-RNN    & 30.55        & 35.47    & 71.70      & 27.31		\\
LSTM        & 32.64        & 34.40    & 72.45      & 25.08      \\
RNN-RBM*     & 33.12        & 34.02    & \textbf{75.40}      & \textbf{28.92}      \\\midrule
LMN-A       & 30.61        & 33.15    & 71.16      & 26.69      \\
LMN-B       & 33.98        & 35.56    & 72.71      & 28.00      \\
pret-LMN-B  & \textbf{34.49}        & \textbf{35.66}    & \textbf{74.16}      & 28.79 \\          
\bottomrule
\end{tabular}
\end{table*}
All the networks have been optimized using Adam~\cite{Kingma2014AdamAM} with a fixed learning rate of $0.001$ using early stopping on the validation set to limit the number of epochs.
Except for the RNN-RBM, all the architectures have a single layer. For the RNN and LSTM models, we have selected the number of hidden recurrent neurons/cells with a grid search over the range $\{50,\ 100,\ 250,\ 500,\ 750\}$. For the LMN architecture, we have searched the number of nonlinear functional units and of the linear memory units over the range $\{(50, 50),\ (50, 100),\ (100, 100),\ (100, 250),\ (250, 250),\ $ $(250, 500)\}$, where the first number refers to functional units and the second to the memory units. All models have been regularized  using L2 weight decay, selecting the regularization hyperparameter by grid search over the range $\{10^{-4},\ 10^{-5},\ 10^{-6},\ 10^{-7},\ 0\}$.

The unrolled network used in pretraining is trained with an unfolding length set to $k=10$, with hidden sizes equal to the corresponding LMN. We found useful to use the SeLU activation function, as defined in \cite{Klambauer2017SelfNormalizingNN}, to improve the convergence of the training procedure for the unfolded model only. Other models, including the final LMN, use a tanh activation function for the hidden units and a sigmoid activation for the outputs. During preliminary experiments we did not find any significant performance improvement when training LMN using different activation functions.

All models are implemented using Pytorch~\cite{paszke2017automatic}.
The test performances for the best configuration of each model (selected on validation) are reported in Table \ref{tbl:midi-results}.

Looking at the results, we notice that the LMN-B model is competitive when confronted with other recurrent architectures with gating units, even without pretraining. The LMN architecture obtains also better results in two different datasets when compared to the RNN-RBM, a more complex architecture which comprises multiple layers. On the Piano MIDI the difference in performance with respect to RNN-RBM is relatively small, while the Nottingham dataset is tested using the updated version, and therefore the results are not exactly comparable. The RNN and LMN performance has also been tested when using a pretraining scheme (note that RNN-RBM uses pretraining as well \cite{BoulangerLewandowski2012ModelingTD}). For the LMN architecture, we notice a more consistent improvement induced by the pretraining algorithm than for the RNN architecture. This is not surprising since the LMN pretraining scheme follows naturally from the equivalence results in \cite{Sperduti2015EquivalenceRB}. LSTM models are not pretrained, and the same pretraining scheme used for RNN and LMN cannot be easily adapted to gating units. We argue that this is an example where it can be clearly appreciated the advantage of dealing with an architecture of lesser complexity which, despite its apparent simplicity, leads to excellent performance results.

\begin{figure}[!ht]
\centering
\includegraphics[width=.35 \textwidth]{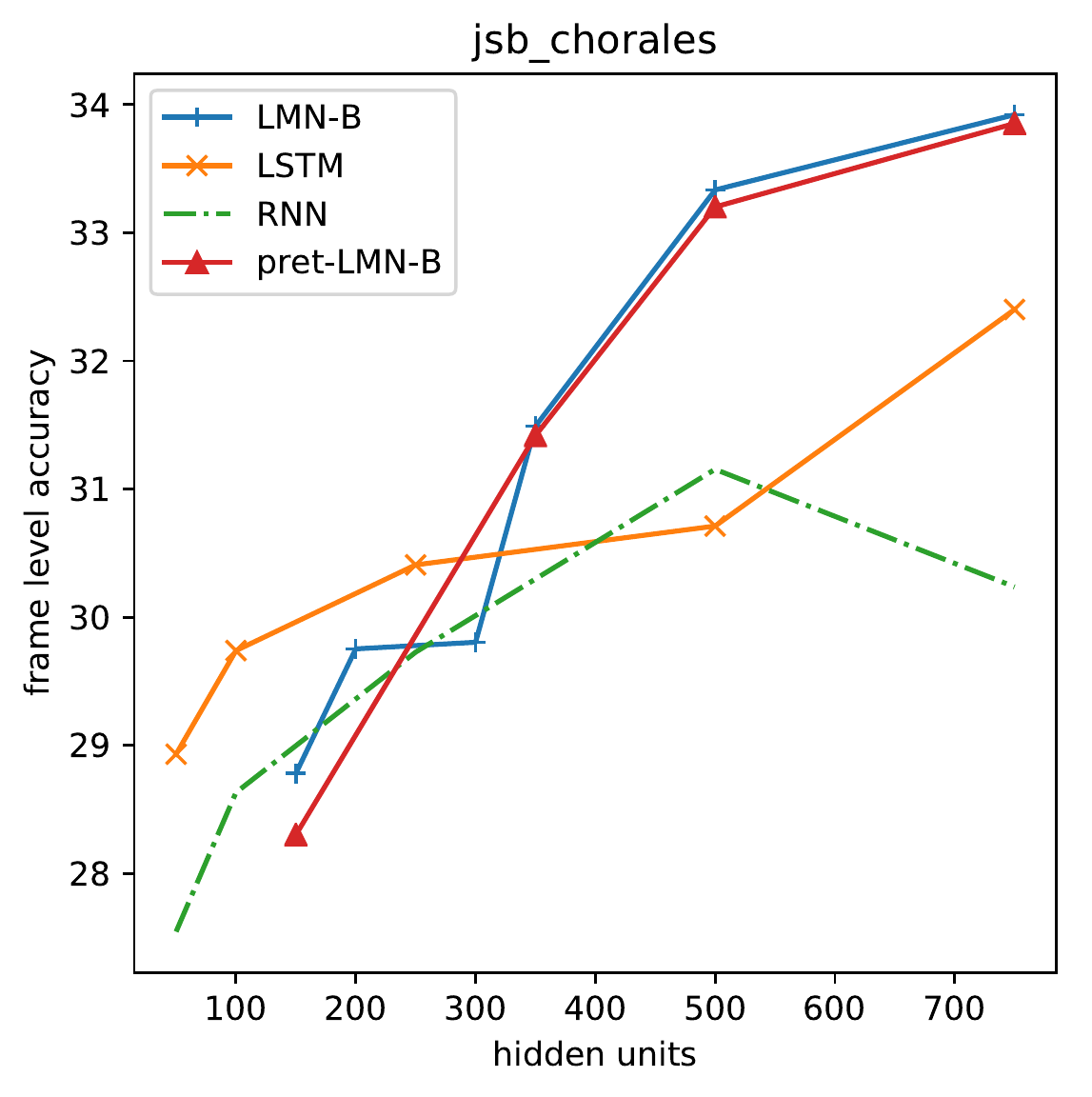}
\caption{Frame level accuracy on the validation set against the number of hidden units for different models trained on the JSB Chorales dataset. For the LMN variants the number of hidden units corresponds to the sum of functional and memory units.}
\label{fig:hid_plots}
\end{figure}

To assess the behavior of the LMN architecture as a function of the parameter space size, Figure~\ref{fig:hid_plots} shows the performance of different models on the validation set for the JSB Chorales. Each curve represents the performance of a model for a given number of hidden units. Again, we focus on the LMN-B architecture both in its basic and pretrained version. For the LMN, we consider the number of hidden units to be the sum of functional and memory units in the configuration under test. We notice a consistent improvement of the LMN models with respect to both LSTM and RNN, starting from 350 hidden units. Please, notice that, as pointed out in Section~\ref{comparison}, with the same number of hidden units LMN-B has significantly less free parameters than the other architectures under comparison.

\begin{figure*}[!ht]
\centering
\resizebox{\textwidth}{!} {
  \begin{subfigure}{.3\textwidth}
    \centering
    \includegraphics[width=.9\linewidth]{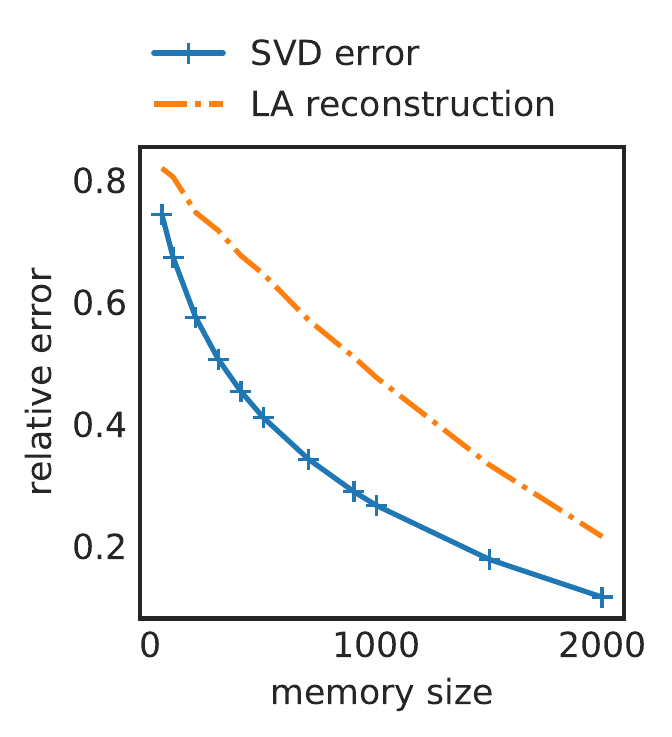}
    \caption{}
    \label{fig:pret-sub1}
  \end{subfigure}%
  \begin{subfigure}{.3\textwidth}
    \centering
    \includegraphics[width=.9\linewidth]{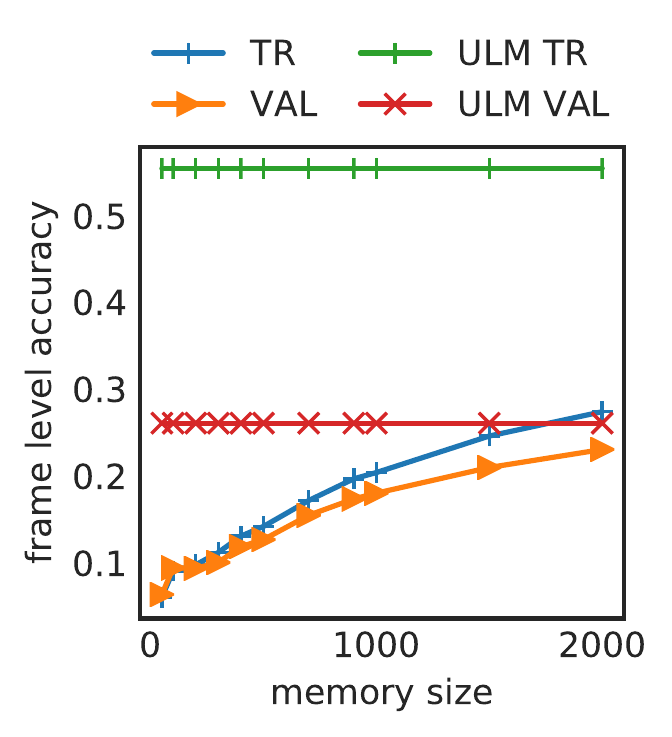}
    \caption{}
    \label{fig:pret-sub2}
  \end{subfigure}
  \begin{subfigure}{.3\textwidth}
    \centering
    \includegraphics[width=.9\linewidth]{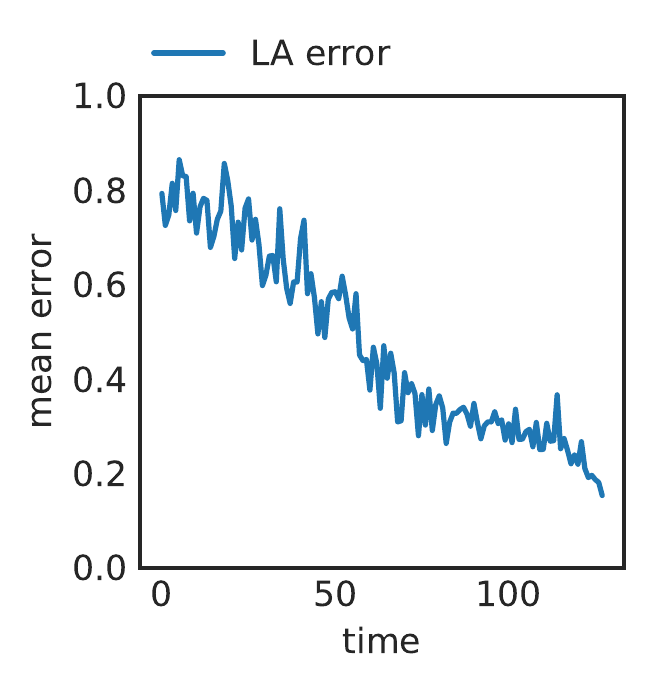}
    \caption{}
    \label{fig:pret-sub3}
  \end{subfigure}
}
\caption{Performance of the pretrained network computed for the JSB Chorales dataset at different stages of the pretraining procedure. Figure \ref{fig:pret-sub1} shows the SVD reconstruction error and the linear autoencoder (LA) reconstruction error for the training set hidden state sequences. Figure \ref{fig:pret-sub2} shows the pretrained LMN error after the initialization, compared against the unfolded network (ULM) on the train (TR) and validation (VAL) sets. Figure \ref{fig:pret-sub3} shows the linear autoencoder reconstruction error for a single hidden state sequence corresponding to the first sample of the training set.}
\label{fig:pretrain}
\end{figure*}

\section{Pretraining Analysis}
In previous sections, we have discussed how, by training a linear autoencoder on the hidden sequences generated by the unfolded network, we can obtain the optimal reconstruction of the hidden states while keeping the minimal amount of hidden units.

To gain a better understanding of the effect of the pretraining procedure, in the following, we study the performance of the model after each step of the algorithm. Figure \ref{fig:pret-sub1} shows the average reconstruction error related to the SVD factorization of $H$, i.e. the matrix containing the hidden states subsequences used to train the corresponding linear autoencoder, for the training set sequences of the JSB Chorales. On the same plot, we overlay the reconstruction error of the corresponding trained linear autoencoder (LA). As expected, the reconstructions error steadily decreases for both models as the number of memory units $h_t^m$ grows. 

The parameter matrices of the linear autoencoder are then used to initialize the LMN according to Eqs. \eqref{eq:pret} to \eqref{eq:pret2}. Figure \ref{fig:pret-sub2} shows the performance obtained by the original unfolded network (ULM) and the corresponding pretrained LMN on the training and validation sets of JSB Chorales. The performance is computed after initialization by pretraining and before the fine-tuning phase. It can be seen how the pretrained-LMN performance on the validation set is close to that obtained by the unfolded network, while LMN greatly reduces the number of parameters used with respect to the ULM configuration. It must also be noted that the performance of the unfolded network is lower than that of a randomly initialized and then trained LMN model: this highlights the need of a fine tuning phase after the pretraining initialization. 

Finally, Figure \ref{fig:pret-sub3} shows the autoencoder reconstruction error for the first training sequence of the JSB Chorales obtained by a pretrained LMN with hidden state size $50$ and $1000$ memory units. It can be noticed that most of the errors are concentrated on the first steps of the sequence, while the second part of the input has a much lower error. This shows that the linear autoencoder needs a burn-in period to recover from the state initialization and after such period becomes fairly accurate.

\section{Conclusion}
We have introduced a novel recurrent architecture, the Linear Memory Network (LMN), designed on a conceptual separation of the memory-related functionalities from the non-linear functional transformations. We build our model on sound theoretical results concerning the equivalence of certain classes of RNN architectures and their unfolded feedforward counterpart. We exploit the same intuition to suggest an effective pretraining scheme to initialize both LMN components in an efficient way. Experimental results show that the model is competitive on difficult tasks against various recurrent architectures and the associated analysis provides insights into the dynamics and properties of the  memory component. We think that the LMN model has the potential of fostering renewed interest in the study of novel, simplified architectures for recurrent neural networks. The availability of a fully linear recurrent dynamics opens up interesting research lines in the direction of efficient training algorithms exploiting closed-form solutions of linear optimization problems. Further, the LMN model can be used as a building block to construct deep or modular architectures or as a replacement of vanilla and gated recurrent units in existing models. Finally, the concept of a recurrent linear memory for sequences can be easily generalized to a recursive autoencoder, allowing to extend the LMN model to the treatment of tree-structured data. 

\bibliographystyle{acm}
\bibliography{biblio}

\end{document}